\documentclass[10pt,twocolumn,letterpaper]{article}

\usepackage[pagenumbers]{wacv} 

\usepackage{graphicx}
\usepackage{amsmath}
\usepackage{amssymb}
\usepackage{booktabs}
\usepackage{balance}

\usepackage{color, colortbl}
\definecolor{PrimaryColumnColor}{rgb}{0.90,0.90,0.90}
\definecolor{SecondaryColumnColor}{rgb}{1,1,1}

\usepackage{makecell}

\usepackage{graphicx}
\usepackage{cite}
\usepackage{amsmath,amssymb,amsfonts}
\usepackage{algorithmic}
\usepackage{graphicx}
\usepackage{textcomp}
\usepackage{soul}
\usepackage{siunitx}
\usepackage{etoolbox,siunitx,booktabs,threeparttable,multirow,array,graphicx}
\usepackage[normalem]{ulem} 
\robustify\bfseries
\robustify\uline
\def\Uline#1{#1\llap{\uline{\phantom{#1}}}}

\usepackage{float}
\usepackage[export]{adjustbox}
\usepackage{xspace}
\usepackage{multicol}
\usepackage{makecell}

\usepackage[pagebackref,breaklinks,colorlinks]{hyperref}
\usepackage[capitalize]{cleveref}
\crefname{section}{Sec.}{Secs.}
\Crefname{section}{Section}{Sections}
\Crefname{table}{Table}{Tables}
\crefname{table}{Tab.}{Tabs.}

\graphicspath{{imgs/}{imgs/old/}{imgs/old/annotation_views/}{imgs/old/bio/}{imgs/old/preprocessing/}{imgs/old/syntetic_vs_our/}}

\def\cit#1{~\cite{#1}}

\usepackage{pifont}
\newcommand{\unet}{U-Net}
\newcommand{\tdunet}{3D \unet}
\newcommand{\nnunet}{nn\unet}
\newcommand{\nnunetres}{\nnunet{} ResEnc}

\newcommand{\swinunet}{Swin\unet{}}
\newcommand{\unetr}{UNETR}
\newcommand{\swinunetr}{Swin-\unetr{}}
\newcommand{\cotr}{CoTr}
\newcommand{\nnformer}{nnFormer}

\newcommand{\levit}{LeViT-UNet-384s}

\newcommand{\missformer}{MISSFormer}
\newcommand{\umamba}{UMamba}
\newcommand{\umambabot}{\umamba Bot}
\newcommand{\umambaenc}{\umamba Enc}
\newcommand{\mednext}{MedNeXt}
\newcommand{\mednextmkt}{\mednext-M-K3}
\newcommand{\mednextmkf}{\mednext-M-K5}
\newcommand{\transunet}{Trans\unet{}}
\newcommand{\transbts}{TransBTS}

\newcommand{\SegMamba}{SegMamba}
\newcommand{\PanSegMamba}{BiSegMamba}
\newcommand{\MultiSegMamba}{MultiSegMamba}
\newcommand{\SegMambaSkip}{SegMambaSkip}

\newcommand\blfootnote[1]{%
  \begingroup
  \renewcommand\thefootnote{}\footnote{#1}%
  \addtocounter{footnote}{-1}%
  \endgroup
}

\def\wacvPaperID{753} 
\def\confName{WACV}
\def\confYear{2025}

\begin{document}

\title{Taming Mambas for Voxel Level 3D Medical Image Segmentation}

\author{
    Luca Lumetti* \qquad
    \,\,\,\,Vittorio Pipoli* \qquad
    Kevin Marchesini \qquad \\
    Elisa Ficarra\,\,\, \qquad 
    Costantino Grana \qquad
    Federico Bolelli \\\\
University of Modena and Reggio Emilia, Italy\\
{\tt\small name.surname@unimore.it}
}

\maketitle
\vspace{-40pt}

\begin{abstract}
Recently, the field of 3D medical segmentation has been dominated by deep learning models employing Convolutional Neural Networks (CNNs) and Transformer-based architectures, each with their distinctive strengths and limitations. CNNs are constrained by a local receptive field, whereas transformers are hindered by their substantial memory requirements as well as they data hungriness, making them not ideal for processing 3D medical volumes at a fine-grained level. For these reasons, fully convolutional neural networks, as \nnunet{}, still dominate the scene when segmenting medical structures in 3D large medical volumes.
Despite numerous advancements towards developing transformer variants with subquadratic time and memory complexity, these models still fall short in content-based reasoning.
A recent breakthrough is Mamba, a Recurrent Neural Network (RNN) based on State Space Models (SSMs) outperforming Transformers in many long-context tasks (million-length sequences) on famous natural language processing and genomic benchmarks while keeping a linear complexity.  
In this paper, we evaluate the effectiveness of Mamba-based architectures in comparison to state-of-the-art convolutional and Transformer-based models for 3D medical image segmentation across three well-established datasets: Synapse Abdomen, MSD Brain Tumor, and ACDC. Additionally, we address the primary limitations of existing Mamba-based architectures by proposing alternative architectural designs, hence improving segmentation performances. The source code is publicly available to ensure reproducibility and facilitate further research.
\footnote{\url{https://anonymous.4open.science/r/WACV2025-TamingMamba/}}
\vspace{-12pt}
\end{abstract}

\section{Introduction}
\label{sec:intro}
\label{sec:introduction}
Image\blfootnote{\hspace{-4pt}*Equal contribution. Authors are allowed to list their name first in their CVs.} segmentation is crucial in the analysis of medical images, typically serving as the preliminary step for examining anatomical structures and surgical planning\cit{asgari2021deep}.
During the latest years Convolutional Neural Networks (CNN)\cit{CNN} and in particular U-shaped Fully Convolutional Neural Networks (FCNN) have known wide adoption in the research community\cit{unet}. Despite their effectiveness, after the outbreak of vision transformers, FCNN has been replaced by hybrid architectures, made up by both Convolutional and Multi-Head Attention layers, aiming to mitigate the problem of local receptive field that characterizes the convolutional operation, relying on the transformer's attention mechanism\cit{transformer}. Several attempts have been made in the literature to integrate transformer-based architectures in the classic U-Net\cit{medformer, chen2021transunet, cao2022swin, hatamizadeh2022unetr, Hatamizadeh2022swin}. Even if these methods led to improvement in performances, it came at the cost of the quadratic memory footprint of the attention mechanism that alongside their data hungriness makes these approaches not ideal when applied to large 3D volumes. 

In such regards, latest research put a lot of efforts in the reduction of the computational cost of the transformer architecture, proposing linear attention mechanism\cit{linformer, performer, longformer} and gating\cit{schlemper2019attention}. Anyway, most of them fall short when it comes to context modeling, in particular when the context-length is considerably high. Recently, the field of sequence modeling has been greatly influenced by an innovative architecture based on State-Space Model (SSM)\cit{SSM1} known as Mamba\cit{mamba}. Mamba shown state-of-the-art capabilities in several Natural Language Processing (NLP) and genomic tasks improving the modeling of big context up to the order of a million tokens, making it a suitable candidate to efficiently process also 3D volumes, where the number of tokens reaches the same order of magnitude.

\noindent
\textbf{Paper Contributions.} 
This paper aims to investigating the effectiveness of Mamba for 3D image segmentation by comparing state-of-the-art Mamba-based architectures with convolutional and Transformer-based segmentation models. Additionally, we seek to address the primary limitations of current Mamba-based architectures by proposing various strategies for integrating Mamba within a \unet-based architecture. Specifically, we examine the impact of modeling directionality on one or more axes and explore the use of Mamba as a selective copying mechanism in skip connections.
To perform our experimental evaluation, we employ three different well know datasets, MSD BrainTumour\cit{Antonelli2022}, Synapse Multi-organ\cit{Landman2015}, and ACDC\cit{Bernard2018}. The code is publicly released to encourage further research.


\section{Related Work}
\label{sec:related}

Convolutional Neural Networks (CNNs)\cit{CNN} have been the dominant solution for both 2D and 3D medical image segmentation for years. Among these, \unet\cit{unet}, characterized by its U-shaped symmetric encoder-decoder structure with skip connections, represents an effective architecture that subsequent models have continued to adopt until the present day. Following \unet, several variants have been introduced, including Res-\unet{}\cit{ResUNet}, Dense-\unet{}\cit{DenseUNet}, V-Net\cit{Milletari2016}, \tdunet{} and its state-of-the-art ecosystem \nnunet{}\cit{isensee2021nnu}, each proposing enhancements to the original framework. Despite their advancements, CNNs inherently face limitations in capturing global patterns due to the locality of the convolutional operator. In response, significant research efforts have been directed towards integrating the attention mechanisms of transformers\cit{transformer} with U-Net-based architectures. This integration aims to leverage both local and global dependencies, as evidenced by models such as MedFormer\cit{medformer}, TransUNet\cit{chen2021transunet}, Swin-UNet\cit{cao2022swin}, UNETR\cit{hatamizadeh2022unetr}, and Swin-UNETR\cit{Hatamizadeh2022swin}. However, the attention mechanism's quadratic complexity forces the imposition of constraints, such as window-based or axial-based attention, to mitigate computational demands. While various studies have attempted to reduce this complexity\cit{reformer,linformer,performer,longformer}, none have matched the performance of traditional attention mechanisms in long-context modeling.

Recent developments have introduced a novel architecture, Mamba\cit{mamba}, predicated on state space modeling\cit{SSM1,SSM2}, which promises capabilities for long-context content-based reasoning with linear-time complexity. Mamba has demonstrated superior performance over state-of-the-art transformer models, such as Pythia-6.9B\cit{biderman2023pythia}, GPT-J-6B\cit{gpt-j}, OPT-6.7B\cit{zhang2022opt}, Hyena\cit{hyena}, in tasks requiring long-context content-based reasoning, such as natural language processing and genomic analyses, with inputs of up to million-length scales. 

Due to their effectiveness and versatility, Mamba-based architectures have been rapidly adapted to various domains, including Computer Vision\cit{vmamba}. In addition, given that segmenting 3D volumes can be seen as processing sequences composed of millions of voxels, several researchers have devoted significant efforts to adapting the Mamba architecture for both 2D and 3D segmentation, yielding promising results\cit{nnMamba,mambaUNet,vmUNet,xing2024segmamba,zhang2024survey}. Among the various contributions, \umamba{} remains one of the most significant in the field given the model ability to adapt effectively to new datasets without the need for extensive hyperparameter tuning. In particular, in their work the authors propose two architectures, namely \umambaenc{} and \umambabot{}, both inheriting their core structure from \unet{} and harnessing Mamba-based layers. The former integrates the Mamba-layers in the encoder part of the architecture, while the latter integrates a single Mamba-layer in the bottleneck. Despite their effectiveness, authors did not focus on the \textit{directionality problem} that derives from employing a recurrent network to extract patterns from data that has more than one spatial dimension. Indeed, once a 3D volume is flattened into a sequence, each voxel is assigned a position within the sequence. This results in the model being able to analyze the latter elements of the sequence by leveraging information from the preceding part. However, it lacks contextual information when processing the initial part of the sequence.

The aforementioned advancements in the field motivated us to devise Mamba architectures for 3D image segmentation, paying particular attention to the directionality. 

\section{Method}
\label{sec:methods}
In this section, all the theoretical concepts related to the vanilla Mamba architecture (a set of stacked Mamba blocks) are introduced. Then, we thoroughly explain how Mamba blocks can be employed to extract patterns from 3D volumes and illustrate approaches to integrate such blocks into a U-Net architecture for 3D medical imaging segmentation. 


\subsection{Preliminaries}
\label{sec:preliminaries}
\noindent \textbf{State Space Models.} A State-Space Model (SSM) is a mathematical representation of a dynamic system which maps a 1D input $x(t) \in \mathbb{R}$ to a $N$D latent state $h(t) \in \mathbb{R}^N$ before projecting it to a 1D output signal $y(t) \in \mathbb{R}$. This system uses $A \in \mathbb{R}^{N \times N}$ as the evolution parameter, $B \in \mathbb{R}^{N \times 1}$, and $C \in \mathbb{R}^{1 \times N}$ as the projection parameters: 
\begin{align}
h'(t) &= Ah(t) + Bx(t) \\
y(t) &= Ch(t) 
\label{eq:cont_SSM}
\end{align}

\noindent Together, the previous equations aim to predict the state of a system from observed data. Since the input is expected to be continuous, the main representation of the SSM is a continuous-time representation.

To employ \cref{eq:cont_SSM} in a real-world scenario, and more specifically into a neural network, a discretization of the variable $t$ is required and can be achieved by introducing a step-size parameter $\Delta$ and a discretization rule, which in this case is the \textit{zero-order hold}:
\begin{align}
    \vspace{-4pt}
    \overline{A} &= \exp(\Delta A) \\
    \overline{B} &= (\Delta A)^{-1}(\exp(\Delta A)-I) \Delta B
    \vspace{-4pt}
\end{align}
This yield to the following discrete state space model that can be computed in a recurrent fashion:
\begin{align}
\vspace{-4pt}
h_{t+1} &= \overline{A}h_t + \overline{B}x_t \\
y_t &= Ch_t
\vspace{-4pt}
\end{align}
This basic SSM performs very poorly in practice due to gradients scaling exponentially in the sequence length. To address this issue, Mamba propose two key elements: imposing a structure to the matrix $A$, using the HiPPO theory\cit{gu2020hippo}, and including a selection mechanism, i.e., making the parameters $B$, $C$, and $\Delta$ input-dependent through a linear projection:

\begin{figure}[bt]
    \centering
    \begin{tabular}{c@{}c@{}}
        \subfloat[]{\includegraphics[width=0.28\textwidth,valign=c]{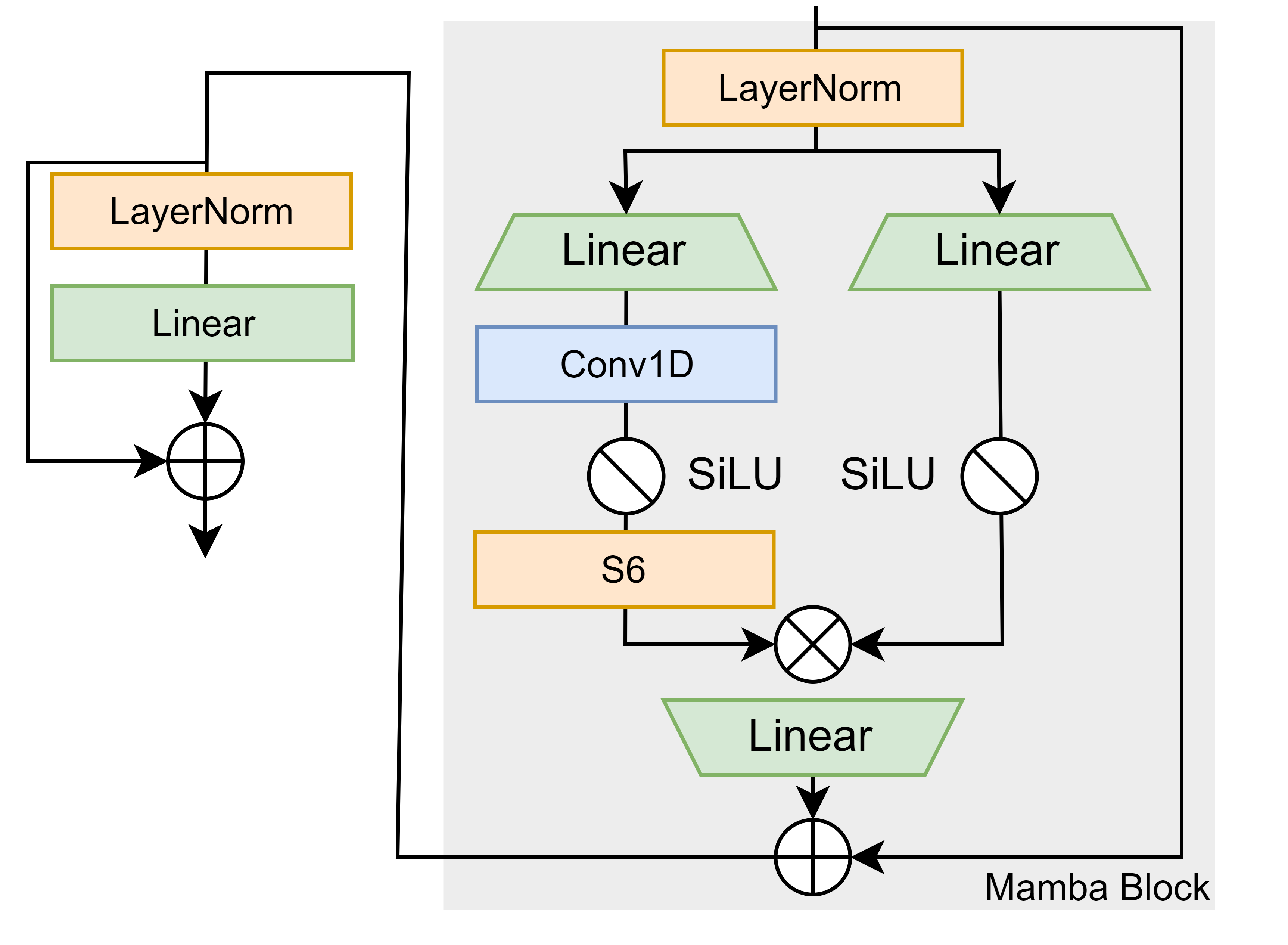}\label{fig:mamba_layer}} & \subfloat[]{\includegraphics[width=0.17\textwidth,valign=c]{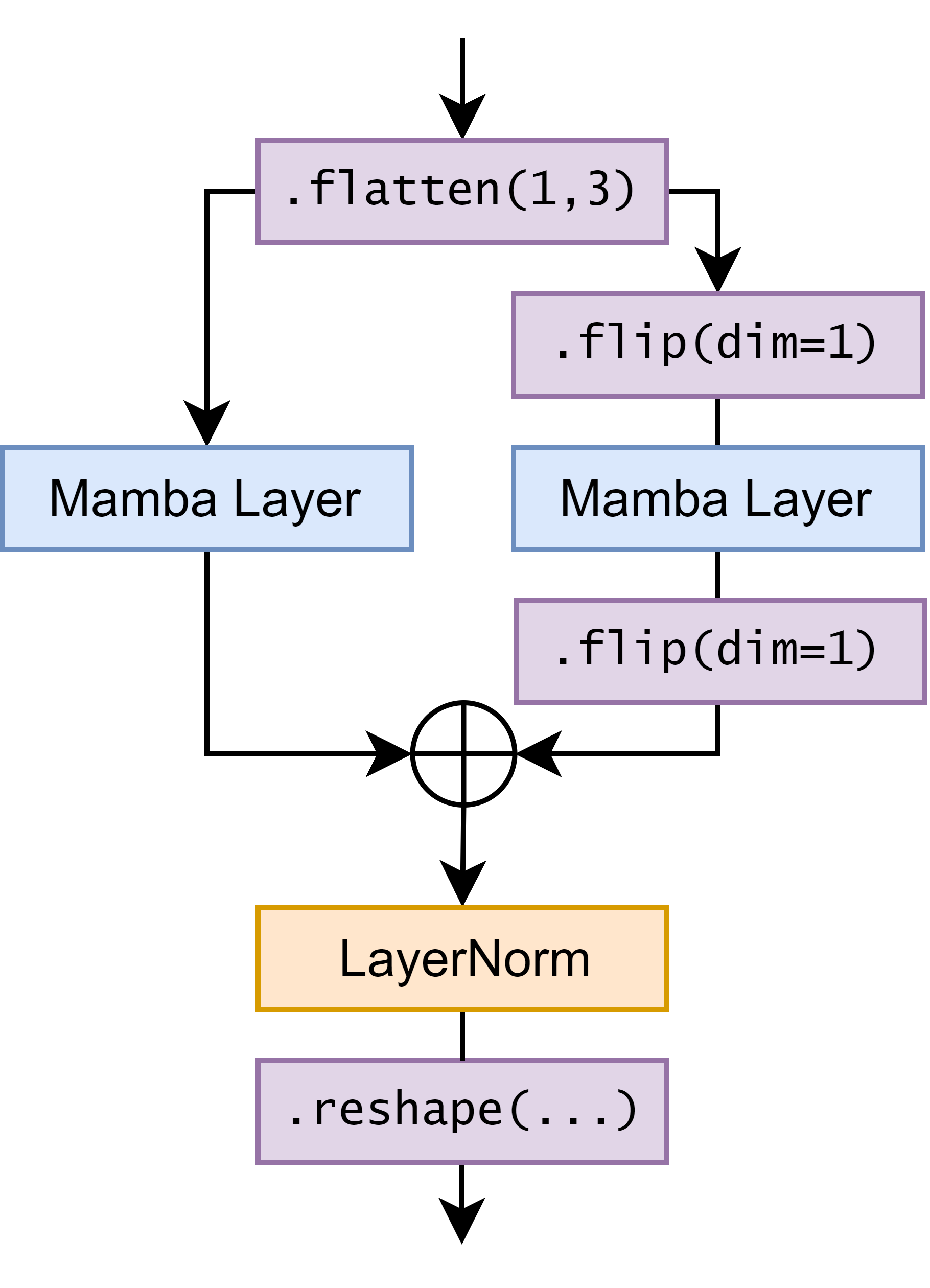}\label{fig:bidir_mamba_layer}}
    \end{tabular}
    \caption{
   From left to right: (a) The unidirectional Mamba layer, which processes input sequences only in the forward direction. The layers within the gray square collectively form the Mamba Block. (b) The bidirectional Mamba layer, consisting of two unidirectional Mamba layers: the left branch processes the forward sequences, while the right branch processes the reversed sequences.
    }
    \label{fig:mamba_block_layer}
    \vspace{-15pt}
\end{figure}


\vspace{-10pt}
\begin{align}
\label{eq:linproj}
    \vspace{-4pt}
    B &= \text{Linear}_N(x) \notag{}\\
    C &= \text{Linear}_N(x) \\
    \Delta &= \text{SoftPlus}(\text{Parameter}+\text{Broadcast}_D(\text{Linear}_1(x)) \notag{}
    \vspace{-4pt}
\end{align}

This formulation, together with an efficient implementation of the process by means of a selective scan algorithm that allows the model to filter out irrelevant information, constitutes the so called \textbf{S6} model\cit{mamba}. To construct a Mamba block, an initial linear projection is employed to expand the input tokens embeddings. Then, a convolution before S6 is applied to prevent independent input token calculations and finally the SSM output is projected back to the original dimension together with a skip connection. SiLU non-linearities are employed\cit{silu}. A visual representation of this block is depicted in gray within \cref{fig:mamba_block_layer}. We refer to the the original publication\cit{mamba} for an in-depth understanding of the discussed methodology.

\begin{figure*}[tb]
    \centering
    \includegraphics[width=0.75\textwidth,valign=c]{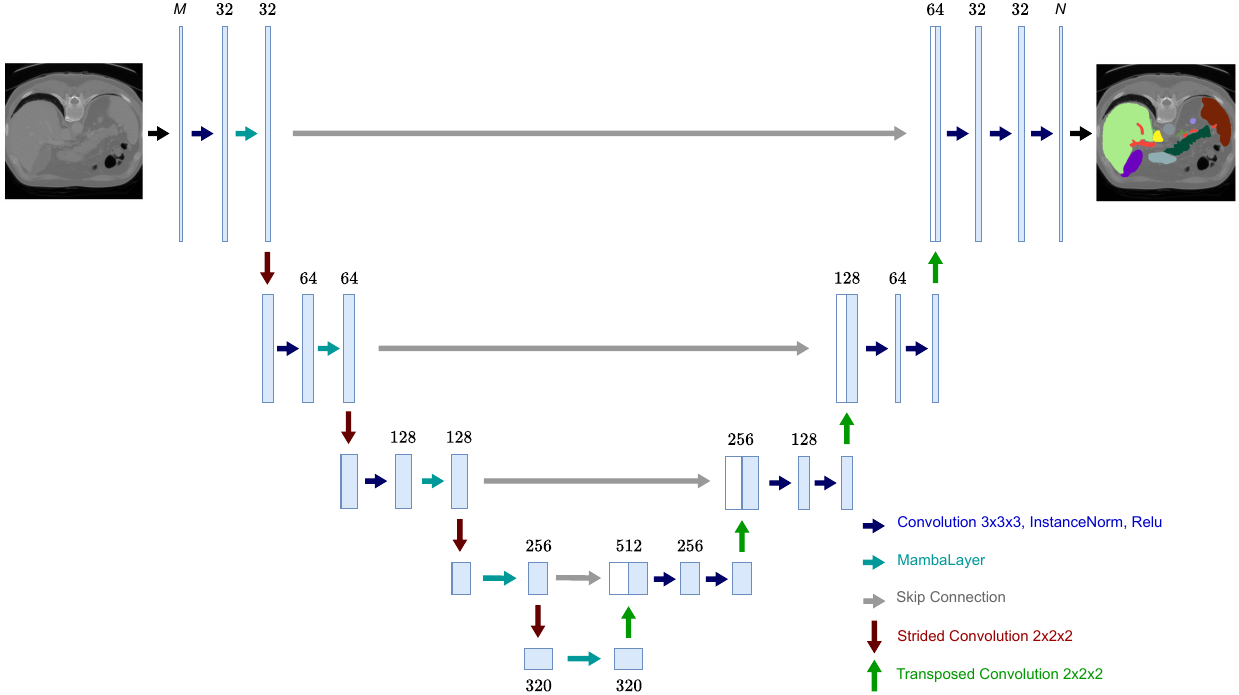} \\
    \caption{
    \unet{} architecture integrating our proposed Mamba Layers. By properly selecting the Mamba Layers (turquoise arrows), \SegMamba{}, \PanSegMamba{}, and \MultiSegMamba{} are obtained. To obtain \SegMambaSkip{} the currently displayed Mamba Layers (turquoise arrows) must be replaced by the standard \unet{} convolution (blue arrow) and corresponding Mamba Layers must be placed within the skip connections (gray arrows).
    }
    \label{fig:model}
\vspace{-15pt}
\end{figure*}

\subsection{Vision Mamba}
Mamba is a sequence to sequence model, thus only able to handle 1D sequences. In order to apply it to 2D images and 3D volumes, a 1D sequence flattenization of pixel (or voxels) is required. Differently from the approach adopted in Vision Transformer, where the quadratic cost of self-attention with respect to the number of pixel prevent their scaling to ``realistic'' input size and requires to extract patches to reduce the input spatial dimension, Mamba allows us a linear-time sequence modelling of the input, preventing any sampling. 
Patch down-sampling is a major issue in medical image segmentation, due to the need for voxel-wise details, which is usually enforced by the large medical input data.

One downside of Mamba is that it is not permutation-invariant. In contrast to the transformer self-attention mechanism, where each token can gather information from every other token in the sequence, Mamba restricts each token to only infer information from the current state, resulting in an approximation of the past tokens only.
This means that when Mamba is employed for image segmentation tasks, the very first pixels (or voxels) in the sequence do not have any context awareness.
For this reason, instead of directly including the Mamba Block into our \unet{} architecture, by taking inspiration from the ViT architecture\cit{dosovitskiy2020image} we developed a wrapper. The wrapping, consisting of an additional LayerNorm and an MLP head followed by a skip connection, allows us to improve Mamba stability. We denote this module as the Mamba Layer, illustrated in \cref{fig:mamba_layer}. Subsequently, we integrated two instances of the Mamba Layer into a unified module. This module, named Bidirectional 3D Mamba Layer, takes as input a 3D volume with dimensions \texttt{(B, H, W, D, C)}. It flattens the spatial dimensions and manages the sequence bidirectionally by feeding one of the two layers with the sequence in the backward direction. Subsequently, the output from this layer is reversed to its original order and then summed token by token with the output of the ``straight'' layer. Finally, the sum is normalized and reshaped back into a 3D volume. This layer is depicted in \cref{fig:bidir_mamba_layer}.

In the following, the strategies we introduce to integrate the Mamba Layer into \nnunet{} are detailed. \cref{fig:model} depicts our \unet{} architectures enriched with Mamba layers. 



\smallskip 
\noindent \textbf{\SegMamba{}.} The initially proposed integration involves the inclusion of a singular (unidirectional) Mamba Layer preceding each pooling convolution and the bottleneck of \unet{}. This strategic placement is designed to enhance the overall contextual understanding, addressing the inherent limitations in global context that convolutions often encounter, while limiting the additional number of parameters.

\smallskip 
\noindent \textbf{\SegMambaSkip{}.} 
One of the universally recognized strength points of the \unet{} architecture lies in its skipped connections\cit{unet}, which allow the decoding part of the network to access fine-grained details coming from the encoder. Meanwhile, Mamba has been devised to efficiently select data in an input-dependent manner, thus being capable of filtering out irrelevant information and remembering relevant ones. Hence, we augment the skip connections in the \unet{} architecture by inserting an additional Bidirectional 3D Mamba Layer before concatenating the activation map to the corresponding decoder output.

\smallskip 
\noindent \textbf{\PanSegMamba{}.} It includes our Bidirectional 3D Mamba Layer before each downsampling step as well as in the \unet{} bottleneck. By leveraging both directions of a single sequential arrangement in \PanSegMamba{}, we strike a balance between computational efficiency and model effectiveness. The Bidirectional 3D Mamba Layer enables the model to effectively weigh the importance of tokens across various spatial dimensions, without the need to consider all possible permutations. This approach is particularly beneficial when dealing with distant dependencies and selective information processing, enhancing the ability to discern relevant features during downsampling and in bottleneck layers.

\begin{figure}[tb]
    \centering
    \includegraphics[width=0.45\textwidth]{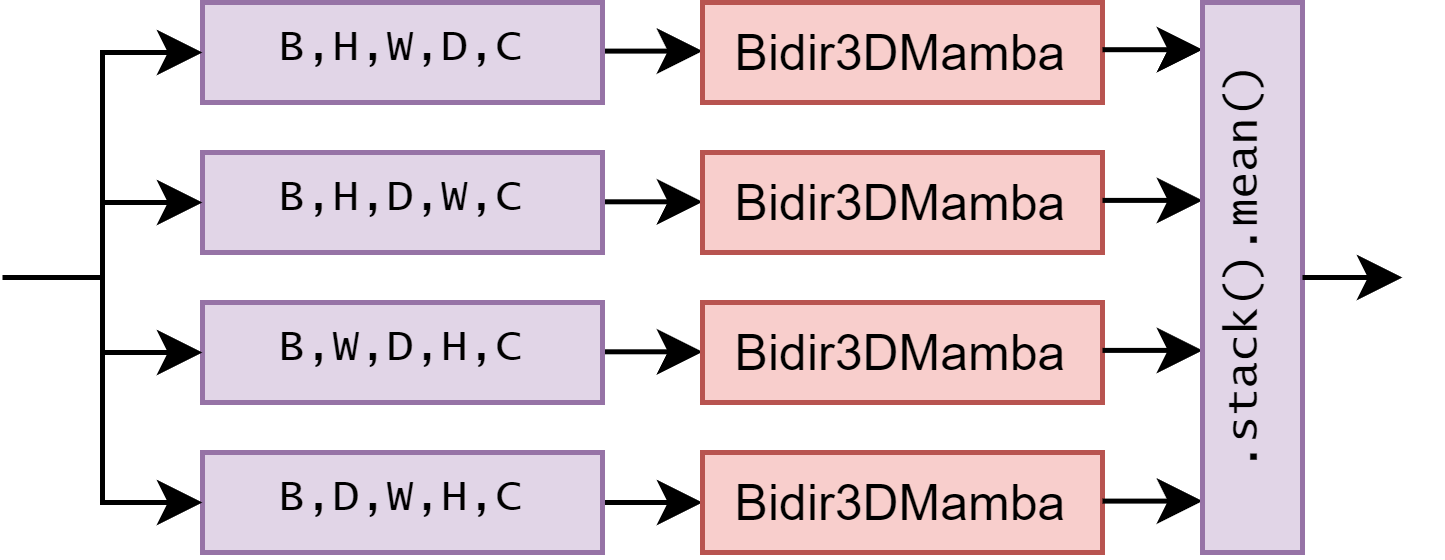}
    \caption{To achieve multi-directionality, four Bidirectional 3D Mamba Layers are employed, corresponding to four out of the six possible permutations of the triplet \texttt{(H, W, D)}. The outputs of each layer is stacked, and the mean per token is computed.}
\label{fig:multidir3dmamba}
\vspace{-15pt}
\end{figure}

\smallskip 
\noindent \textbf{\MultiSegMamba{}.}
The order of input tokens is important within Mamba. For this reason, we propose to process all conceivable sequential arrangements of the volume, resulting in a total of six possible permutations for the three spatial dimensions \texttt{(H, W, D)} of a 3D volume. This yields a total of $12$ distinct sequences, accounting for both the forward and backward directions of the six permutations.
The rationale behind seeking multiple directions stems from the necessity for each voxel to exploit spatial information in all conceivable orientations. If we were to consider only a single sequence, such as \texttt{(H, W, D).flatten()}, the distance between the first token at index \texttt{(0, 0, 0)} and the token at index \texttt{(0, 0, 1)} would be $H*W$ instead of $1$. Typically, the values of $H$ and $W$ are in the order of $10^2$, resulting in a total distance of $10^4$. Due to memory constraints, we only encompass 4 out of 6 possible directions.\footnote{We use \texttt{(H, W, D)}, \texttt{(H, D, W)}, \texttt{(W, H, D)}, and \texttt{(D, W, H)}.} By incorporating multiple directions, we maintain linear complexity while affording each token superior spatial awareness. This approach ensures that neighboring tokens are indeed proximate in the obtained representation, enhancing the overall spatial awareness of the model. To aggregate the output sequences of all the modules involved, we stack each sequence on a new axis and compute the mean value across it (\cref{fig:multidir3dmamba}). This module substitute the Bidirectional 3D Mamba Layer in \PanSegMamba{}.

\subsection{Implementation Details}
Details regarding patch shape, batch size, and other pipeline settings are reported in \cref{tab:config}. All of our models have been trained for $300$ epochs using RAdam, a learning rate of $0.0003$ and a linear learning rate scheduler. For the parameters initialization of the Mamba layers, we scale the weights of residual layers at initialization by a factor of $1/\sqrt{N}$ where $N$ is the number of residual layers. This is the same as in GPT-2 paper and employed in the Mamba source code.
The inner dimension of the State Space Model (i.e., the size of the evolution parameter $A$, \cref{sec:preliminaries}) is defined as $\min(C, 256)$ where $C$ is the number of token's channels in the input sequence. Training has been performed on an A100 Nvidia GPU using CUDA 11.8 and PyTorch 2.1.2.

\section{Experiments and Results}
\label{sec:results}

\smallskip
\noindent \textbf{Datasets.} We conducted experiments using three different well known datasets: MSD BrainTumour\cit{Antonelli2022}, Synapse Multi-organ\cit{Landman2015}, and ACDC\cit{Bernard2018}. 
It is worth noticing that the selected experimental setting aligns with the existing literature on medical image segmentation\cit{zhou2023nnformer,Shaker2024,xu2023levit,cao2022swin,hatamizadeh2022unetr,xie2021cotr}

\textit{MSD BrainTumour.} The first one is the BrainTumour segmentation dataset from the Medical Segmentation Decathlon (MSD BrainTumour)\cit{Antonelli2022}. It consists of 484 MRI images, each containing four channels: FLAIR, T1w, T1gd, and T2w. The images were annotated with three tumour sub-regions: edema (ED), enhancing tumour (ET), and non-enhancing tumour (NET). To be coherent with the results reported by\cit{hatamizadeh2022unetr}, the segmentation metrics were computed on the classes ET, tumour core (TC, which is the union of ET and NET), and whole tumour (WT, which is the union of ED, ET and NET). Following the split provided by\cit{hatamizadeh2022unetr}, we employed 95\% of the dataset as a training/validation set with 5-fold cross-validation, and 5\% for testing purposes.

\begin{table}[t]
  \setlength{\tabcolsep}{6pt}
  \footnotesize
  \centering
  \caption{Configuration of the proposed models when trained on the selected datasets.}
  \setlength{\tabcolsep}{2pt} 
  \begin{tabular}{l
  >{\columncolor{PrimaryColumnColor}}c
  c
  >{\columncolor{PrimaryColumnColor}}c
  }
    \toprule
    & \cellcolor{SecondaryColumnColor} \textbf{BrainTumour} & \cellcolor{SecondaryColumnColor} \textbf{Synapse} & \cellcolor{SecondaryColumnColor} \textbf{ACDC} \\
    \midrule
    Spacing & [1, 1, 1] & [3, 0.76, 0.76] & [6.35, 1.52, 1.52] \\
    Median shape & $138 \times 170 \times 138$ & $148 \times 512 \times 512$ & $13 \times 246 \times 213$ \\
    Crop size    & $128 \times 128 \times 128$ & $48 \times 192 \times 192$ & $14 \times 256 \times 224$  \\
    Batch size   & 2 & 2 & 4 \\
    \bottomrule
  \end{tabular}
  \label{tab:config}
  \vspace{-10pt}
\end{table}

\newcolumntype{g}{>{\columncolor{LightCyan}}S[table-auto-round,table-format=2.2]}
\newcolumntype{a}{>{\columncolor{white}}S[table-auto-round,table-format=2.2]}

\begin{table*}[tb]
  \setlength{\tabcolsep}{6pt}
  \caption{5-fold cross-validation results on the BrainTumor dataset. Our proposals are marked with $\dagger$. Standard deviations for the average scores over the 5 folds are reported. Best results are in \textbf{bold} while the second best are \underline{underlined}.}
  \footnotesize
  \centering
  
  \begin{tabular}{c l
  >{\columncolor{PrimaryColumnColor}}S[table-auto-round,table-format=2.2(2),separate-uncertainty=true]
  >{\columncolor{PrimaryColumnColor}}S[table-auto-round,table-format=2.2(2),separate-uncertainty=true]
  >{\columncolor{SecondaryColumnColor}}S[table-auto-round,table-format=2.2]
  >{\columncolor{SecondaryColumnColor}}S[table-auto-round,table-format=2.2]
  >{\columncolor{PrimaryColumnColor}}S[table-auto-round,table-format=2.2]
  >{\columncolor{PrimaryColumnColor}}S[table-auto-round,table-format=2.2]
  >{\columncolor{SecondaryColumnColor}}S[table-auto-round,table-format=2.2]
  >{\columncolor{SecondaryColumnColor}}S[table-auto-round,table-format=2.2]}
    \toprule
    &\multicolumn{1}{c}{\multirow{2}{*}[0pt]{\textbf{Model}}} & \multicolumn{2}{>{\columncolor{SecondaryColumnColor}}c}{\textbf{Average}} & \multicolumn{2}{>{\columncolor{SecondaryColumnColor}}c}{\textbf{WT}} & \multicolumn{2}{>{\columncolor{SecondaryColumnColor}}c}{\textbf{ET}} & \multicolumn{2}{>{\columncolor{SecondaryColumnColor}}c}{\textbf{TC}} \\
    && \cellcolor{SecondaryColumnColor} \textbf{HD95↓} & \cellcolor{SecondaryColumnColor} \textbf{DSC↑} & \textbf{HD95↓} & \textbf{DSC↑} & \cellcolor{SecondaryColumnColor} \textbf{HD95↓} & \cellcolor{SecondaryColumnColor} \textbf{DSC↑} & \textbf{HD95↓} & \textbf{DSC↑} \\
    \midrule
    \multirow{4}{*}{\rotatebox{90}{CNNs}} & \nnunet{} \cite{isensee2021nnu} &4.53 (017)&85.74 (093)&4.21&91.15&3.89&80.76&5.47&85.29\\
    & \nnunetres{} \cite{isensee2021nnu} &4.12(016)&85.6(0.70)&3.71&89.93&\Uline{3.72}&80.86&4.93&86.01\\
    & \mednextmkt{} \cite{Roy2023} &6.35(021)&85.27(045)&4.59&90.84&6.57&\Uline{80.88}&7.89&84.1\\
    & \mednextmkf{} \cite{Roy2023} &6.67(030)&84.79(081)&4.93&88.93&6.75&79.97&8.33&85.47\\
    \midrule
    \multirow{9}{*}{\rotatebox{90}{Transformers}} & \transunet{} \cite{chen2021transunet} &13.18(083)&64.14(084)&14.42&70.16&10.8&54.31&14.31&67.94\\
    & \transbts{} \cite{wang2021transbts} &9.83(022)&69.72(056)&10.32&78.22&10.2&57.26&8.97&73.68\\
    & \cotr{}    \cite{xie2021cotr} &9.96(019)&68.21(042)&9.25&74.81&9.58&55.14&11.04&74.67\\
    & \unetr{}   \cite{hatamizadeh2022unetr} &9.04(041)&70.92(102)&8.03&78.85&9.83&58.06&9.25&75.85\\
    & \swinunet{} \cite{cao2022swin} &9.98(033)&67.95(057)&8.85&76.43&10.31&57.21&10.77&70.2\\
    & \swinunetr{} \cite{Hatamizadeh2022swin} &6.77(025)&84.07(067)&7.13&88.92&7.54&79.93&5.63&83.34\\
    & \levit{} \cite{xu2023levit} &8.56(030)&70.06(087)&8.2&77.06&8.6&58.1&8.89&75.03\\
    & \missformer{} \cite{huang2021missformer} &9.21(019)&83.08(039)&8.4&88.21&9.57&79.71&9.64&81.33\\
    & \nnformer{} \cite{zhou2023nnformer} &4.05(025)&86.34(051)& \Uline{3.47}&91.28&4.24&\bfseries81.76&4.46&85.97\\
    \midrule
    \multirow{6}{*}{\rotatebox{90}{Mamba}} & \umambabot \cite{Ma2024} &\bfseries 3.80(022)&86.35(077)&3.49&92.1&3.8&80.04&4.1&86.9\\
    & \umambaenc \cite{Ma2024} &4.17(018)&86.16(053)&3.63&\Uline{92.30}&4.44&79.72&4.43&86.46\\
    & \SegMambaSkip{}$\dagger$ &4.53(020)&85.25(081)&3.61&92.11&5.43&78.85&4.54&84.79\\
    & \SegMamba{}$\dagger$     &\Uline{3.82(011)}&\Uline{86.66(045)}&3.66&92.26&3.83&80.77&\Uline{3.96}&\Uline{86.96}\\
    & \PanSegMamba{}$\dagger$  &3.85(016)&85.75(0.60)&\bfseries 3.38&\bfseries 92.43&\bfseries3.46&79.6&4.7&85.21\\
    & \MultiSegMamba{}$\dagger$ &3.84(024)&\bfseries 86.7(0.51)&3.72&92.09&3.88&80.84&\bfseries3.93&\bfseries87.18\\
    \bottomrule
  \end{tabular}
  \label{tab:brain_results}
    \vspace{0pt}
\end{table*}

\textit{Synapse Multi-organ.} The second dataset is the Synapse Multi-organ segmentation da\-ta\-set\cit{Landman2015}, published within the MICCAI 2015 Multi-Atlas Abdomen Labeling Challenge. This dataset includes $3\,779$ axial contrast-enhanced abdominal CT images from 30 abdominal CT scans, with each volume consisting of 85 to 198 slices. We adopted the same split as in\cit{chen2021transunet}, with 18 cases for training and 12 cases for testing. In line with our competitors, the evaluation metrics for this dataset were calculated for eight out of thirteen annotated abdominal organs: aorta, gallbladder, left kidney, right kidney, liver, pancreas, spleen and stomach.

\textit{ACDC.} Lastly, the third dataset employed is the ACDC dataset\cit{Bernard2018}. It comprises 100 MRI scans, each labeled for the left ventricle (LV), right ventricle (RV), and myocardium (Myo). We divided this dataset into 80 samples for training and validation, and 20 test samples, following the split described in\cit{chen2021transunet}. 

\smallskip
\noindent \textbf{Evaluation Metrics.} We employ Dice Similarity Coefficient (DSC in \%) and the 95th percentile Hausdorff Distance (HD95 in mm), two widely accepted metrics for segmentation task\cit{Maier2024}. 


The DSC has practically the same meaning as the IoU (Intersection over Union), but the first one is better suited when the region of interest is much smaller than the background. In such a scenario, DSC can be more robust and informative than IoU since more weight is given to the correctly identified region. The DSC metric, and its relationship with the IoU, are expressed by the following formula:
\begin{equation}
    \text{DSC(P,GT)} = \frac{2 \times |P \cap GT|}{|P| + |GT|} =  \frac{2 \times IoU}{1 + IoU}
\end{equation}

\noindent where $P$ is the model prediction and $GT$ is the ground truth.

On the other hand, the HD95 computes the maximum distance between two sets of points, considering the 95th percentile of these distances. In general, the 95th percentile of the distances between boundary points in A and B is defined as follows:
\begin{equation}
    \text{d}_{95}(A,B) = \text{x}^{95}_{a \in A}\left\{\min_{b \in B} d(a, b)\right\}
\end{equation}
where ${x}_{a \in A}^{95}\{\}$ denotes the 95th percentile of the elements in the set enclosed within the brackets.
Given the set formed by the pixels in the predicted mask ($P$) and the set of pixels belonging to the ground truth ($GT$), the Hausdorff distance is determined as the maximum value of the two distances between $P$ and $GT$ and $GT$ and $P$ at the 95th percentile:
\begin{equation}
    \text{HD95(P,GT)} = \max\biggl\{d_{95}(P,GT), d_{95}(GT,P)\biggr\}
\end{equation}
By using the 95th percentile, this metric provides a robust evaluation that is less sensitive to outliers or extreme differences between the sets of points.

\begin{table*}[tb]
  \setlength{\tabcolsep}{6pt}
  \footnotesize
  \centering
  \caption{5-fold cross-validation results on the Synapse Abdomen dataset. Our proposals are marked with $\dagger$. For space constraints, single class results only report the Dice score. Standard deviations for the average scores over the 5 folds are reported.
  Best results are in \textbf{bold} while the second best are \underline{underlined}.}
  \begin{tabular}{c l
  >{\columncolor{PrimaryColumnColor}}S[table-auto-round,table-format=2.2(2),separate-uncertainty=true]
  >{\columncolor{PrimaryColumnColor}}S[table-auto-round,table-format=2.2(2),separate-uncertainty=true]
  c
  >{\columncolor{PrimaryColumnColor}}S[table-auto-round,table-format=2.2]
  c
  >{\columncolor{PrimaryColumnColor}}S[table-auto-round,table-format=2.2]
  c
  >{\columncolor{PrimaryColumnColor}}S[table-auto-round,table-format=2.2]
  c
  >{\columncolor{PrimaryColumnColor}}S[table-auto-round,table-format=2.2]}
    \toprule
    & \multicolumn{1}{c}{\multirow{3}{*}[0pt]{\textbf{Model}\vspace{-6pt}}} & \cellcolor{SecondaryColumnColor} & \cellcolor{SecondaryColumnColor} & & \cellcolor{SecondaryColumnColor} &  \\
    & & \multicolumn{2}{>{\columncolor{SecondaryColumnColor}}c}{\multirow{-2}{*}[0pt]{\textbf{Average}}} & &  \cellcolor{SecondaryColumnColor} &  \\
    \cmidrule(l){3-4}
    & & \cellcolor{SecondaryColumnColor} \textbf{HD95↓} & \cellcolor{SecondaryColumnColor} \textbf{DSC↑} & 
    \multirow{-3}{*}[-2pt]{\rotatebox[origin=c]{90}{\textbf{Aorta}}}  &
    \multicolumn{1}{>{\columncolor{SecondaryColumnColor}}c}{\multirow{-3}{*}[-2pt]{\rotatebox[origin=c]{90}{\textbf{Gallb.}}}} &
    \multicolumn{1}{>{\columncolor{SecondaryColumnColor}}c}{\multirow{-3}{*}[3pt]{\rotatebox[origin=c]{90}{\textbf{L.Kidn.}}}} &
    \multicolumn{1}{>{\columncolor{SecondaryColumnColor}}c}{\multirow{-3}{*}[4pt]{\rotatebox[origin=c]{90}{\textbf{R.Kidn.}}}} &
    \multicolumn{1}{>{\columncolor{SecondaryColumnColor}}c}{\multirow{-3}{*}[-4pt]{\rotatebox[origin=c]{90}{\textbf{Liver}}}}  & 
    \multicolumn{1}{>{\columncolor{SecondaryColumnColor}}c}{\multirow{-3}{*}[-1pt]{\rotatebox[origin=c]{90}{\textbf{Pancr.}}}} & 
    \multicolumn{1}{>{\columncolor{SecondaryColumnColor}}c}{\multirow{-3}{*}[-1pt]{\rotatebox[origin=c]{90}{\textbf{Spleen}}}} & 
    \multicolumn{1}{>{\columncolor{SecondaryColumnColor}}c}{\multirow{-3}{*}[-2pt]{\rotatebox[origin=c]{90}{\textbf{Stom.}}}}  \\
    \midrule
    \multirow{4}{*}{\rotatebox{90}{CNNs}} &\nnunet{} \cite{isensee2021nnu} &10.91(0.69)&86.21(1.19)&91.65&70.01&86.67&85.75&96.11&83.22&90.69&85.55\\
    & \nnunetres{} \cite{isensee2021nnu} &7.70(042)&86.61(1.07)&89.94&64.2&90.79&91.18&\bfseries 97.36&79.48&92.03&\bfseries87.93\\
    & \mednextmkt{} \cite{Roy2023} &18.99(053)&85.7(0.77)&\Uline{92.44}&\bfseries 72.75&87.62&86.21&\Uline{97.15}&81.17&90.3&77.93\\
    & \mednextmkf{} \cite{Roy2023} &17.30(060)&86.(0.86)&92.15&71.66&87.89&87.43&96.91&80.26&90.95&80.78\\
    \midrule
    \multirow{9}{*}{\rotatebox{90}{Transformers}}  & \transunet{} \cite{chen2021transunet}         &32.27(1.01)&77.24(0.91)&86.88&62.59&81.35&76.98&94.45&55.57&84.97&75.12\\
    & \transbts{} \cite{wang2021transbts} &11.98(0.67)&83.27(1.06)&91.95&62.24&86.91&87.15&96.67&71.91&91.62&77.7\\
    & \cotr{} \cite{xie2021cotr} &9.35(039)&84.67(0.75)&\bfseries 92.77&63.07&87.98&86.84&92.75&78.63&94.54&80.76\\
    & \unetr{} \cite{hatamizadeh2022unetr}  &19.15(084)&78.1(1.12)&89.75&55.81&85.71&84.71&94.00&60.23&84.47&70.14\\
    & \swinunet{} \cite{cao2022swin} &22.02(070)&79.06(0.73)&85.65&66.46&83.03&79.37&94.02&56.57&90.67&76.72\\
    & \swinunetr{} \cite{Hatamizadeh2022swin} &11.02(072)&83.64(1.31)&91.22&66.48&87.09&86.62&95.99&68.79&95.72&77.19\\
    & \levit{} \cite{xu2023levit}    &16.80(081)&78.38(0.99)&87.52&61.77&84.04&79.87&92.8&59.2&88.84&73.03\\
    & \missformer{} \cite{huang2021missformer} &18.50(059)&81.87(0.85)&86.48&68.92&85.56&81.6&94.24&65.44&91.7&80.99\\
    & \nnformer \cite{zhou2023nnformer} &11.14(048)&86.56(0.64)&91.63&69.85&86.61&86.55&96.97&\bfseries 83.68&90.72&86.44\\
    \midrule
    \multirow{6}{*}{\rotatebox{90}{Mamba}} &\umambabot{}  \cite{Ma2024}  &7.35(042)&86.88(0.80)&89.88&60.14&89.99&94.37&96.81&82.33&95.66&85.88\\
    &\umambaenc{}  \cite{Ma2024}  &7.83(050)&87.82(0.75)&89.57&65.2&89.46&94.84&96.97&\Uline{83.35}&\bfseries96.80&86.4\\
    &\SegMambaSkip{}$\dagger$ &6.29(047)&88.26(0.89)&89.64&69.04&93.40& 94.91&\bfseries96.80&79.61&\Uline{96.45}&86.19\\
    &\SegMamba{}$\dagger$ &7.91(038)&87.48(0.77)&89.59&62.21&\Uline{93.65}&94.81&96.82&80.72&95.22&\Uline{86.85}\\
    &\PanSegMamba{}$\dagger$ &\Uline{ 5.99(053)}&\Uline{88.29(0.90)}&91.02&70.12&92.98&94.32&96.94&79.08&96.26&85.58\\
    &\MultiSegMamba{}$\dagger$ &\bfseries 5.98(036)&  \bfseries88.93(0.84)&91.36&\Uline{71.78}&\bfseries 94.00&\Uline{94.88}&95.76&80.65&96.22&86.77\\
    \bottomrule
  \end{tabular}
  \label{tab:synapse_results}
  \vspace{-10pt}
\end{table*}

\begin{table}[tb]
  \setlength{\tabcolsep}{6pt}
  \footnotesize
  \centering
  \caption{5-fold cross-validation results on the Automatic Cardiac Diagnosis (ACDC) dataset. Our proposals are marked with $\dagger$. The evaluation metric is the DSC (\%). Best results are in bold while second best are underlined.}
  \begin{tabular}{c l
  >{\columncolor{PrimaryColumnColor}}S[table-auto-round,table-format=2.2]
  c
  >{\columncolor{PrimaryColumnColor}}S[table-auto-round,table-format=2.2]
  c
  }
    \toprule
    & \textbf{Model} & \cellcolor{SecondaryColumnColor} \textbf{Average} & \textbf{RV} & \cellcolor{SecondaryColumnColor} \textbf{Myo}   & \textbf{LV}    \\
    \midrule
    \multirow{4}{*}{\rotatebox{90}{CNNs}} & \nnunet{} \cite{isensee2021nnu}  &91.42&90.1&88.74&95.41\\
    & \nnunetres \cite{isensee2021nnu} &90.84&89.17&88.52&94.84\\
    & \mednextmkt{} \cite{Roy2023} &91.64&89.43&\Uline{89.77}&95.72\\
    & \mednextmkf{} \cite{Roy2023} &90.7&88.5&88.88&94.73\\
    \midrule
    \multirow{9}{*}{\rotatebox{90}{Transformers}} & TransUNet \cite{chen2021transunet} &89.75&88.88&84.66&95.7\\
    & \transbts \cite{wang2021transbts} &91.29&90.42&87.94&95.51\\
    & \cotr \cite{xie2021cotr} &90.9&89.17&88.34&95.18\\
    & \unetr{} \cite{hatamizadeh2022unetr} &88.72&85.55&86.48&94.12\\
    & \swinunet{} \cite{cao2022swin} &89.97&88.29&85.61&\bfseries96.01\\
    & \swinunetr{} \cite{Hatamizadeh2022swin} &91.36&\Uline{90.48}&87.84&\Uline{95.75}\\
    & \levit{} \cite{xu2023levit} &90.21&89.78&87.1&93.75\\
    & \missformer{} \cite{huang2021missformer} &87.73&86.55&85.24&91.42\\
    & \nnformer{} \cite{zhou2023nnformer}    &\Uline{91.87}&\bfseries90.78&89.37&95.46\\
    \midrule
    \multirow{6}{*}{\rotatebox{90}{Mamba}}
    & \umambabot{} \cite{Ma2024} &90.44&87.67&88.76&94.89\\
    & \umambaenc{} \cite{Ma2024} &90.07&87.34&88.23&94.65\\
    & \SegMambaSkip{}$\dagger$ &91.49&89.58&89.51&95.39\\
    & \SegMamba{}$\dagger$  &91.33&89.37&89.4&95.22\\
    & \PanSegMamba{}$\dagger$  &91.5&89.46&89.66&95.37\\
    & \MultiSegMamba{}$\dagger$ &\bfseries 92.04&90.39&\bfseries90.29&95.44\\
    \bottomrule
  \end{tabular}
  \label{tab:acdc_results}
  \vspace{-0pt}
\end{table}


\smallskip
\noindent \textbf{Compared Methods.}
Performance comparison has been performed on recently proposed methods for medical image segmentation. Specifically, considered competitors can be classified into three main groups: CNN-, Transformer-, and Mamba-based architectures. 

In the former group, we include the original \nnunet{}\cit{isensee2021nnu} configuration making use of the vanilla \unet{} architecture (\nnunet{}), and its variations based on the \unet{} with residual connections in the encoder (\nnunetres{}). Furthermore, the transformer-inspired-CNN-modification based on ConvNeXt blocks, MedNeXt\cit{Roy2023}, has been considered in its two variations K3, and K5. 
For what concerns Tranformer-based architectures, we compare our proposals with \transunet{}\cit{chen2021transunet}, \transbts{}\cit{wang2021transbts}, \cotr{}\cit{xie2021cotr}, an hybrid architecture combining convolutional and transformer modules, \unetr{}\cit{hatamizadeh2022unetr}, \swinunet{}\cit{cao2022swin} and its \unetr{}-based variation \swinunetr{}\cit{Hatamizadeh2022swin}, \levit{}\cit{xu2023levit}, \missformer{}\cit{huang2021missformer}, and the recently published \nnformer{}\cit{zhou2023nnformer}.
Finally, we include \umamba{}\cit{Ma2024} in its two variations \umambabot{} and \umambaenc{}. 


In our experiments, a standardized scheme for hyperparameter configuration has been adopted. Whenever available, the self-configuration method capabilities have been employed, otherwise, we opted for the default configuration (if any) or the one closest to the respective dataset, reducing the learning rate until convergence. Models are trained from scratch without any pre-training data. The \nnunet{} five-fold cross-validation schema has always been employed.

\begin{table}[tb]
  \setlength{\tabcolsep}{3pt}
  \footnotesize
  \centering
  \caption{Computational comparison on the Synapse dataset. Our proposals are marked with $\dagger$. The number of parameters is expressed in millions [M] and VRAM in gigabyte [GB]. Training and inference times, expressed in hours [h] and seconds [s], respectively, are obtained on an Nvidia A100 with 80GB of memory. All competitor models were trained for 1000 epochs, as recommended by most of their original papers, while our method achieved convergence in only 300 epochs. Inference times is the average across all test volumes.}
  \begin{tabular}{c l
  >{\columncolor{PrimaryColumnColor}}S[table-auto-round,table-format=2.2]
  c
  >{\columncolor{PrimaryColumnColor}}S[table-auto-round,table-format=2.2]
   c
   >{\columncolor{PrimaryColumnColor}}S[table-auto-round,table-format=2.2]
  }
  
    \toprule
    & \textbf{Models} & \cellcolor{SecondaryColumnColor} \textbf{Params} & \textbf{GFLOPs} & \cellcolor{SecondaryColumnColor} \textbf{VRAM} & \textbf{Tr.}  &  \cellcolor{SecondaryColumnColor} \textbf{Inf.} \\
    \midrule
    \multirow{4}{*}{\rotatebox{90}{CNNs}} & \nnunet{} \cite{isensee2021nnu}  & 30.64 & 410.11 & 7.65 & 9.2 & 21.8 \\ 
    & \nnunetres \cite{isensee2021nnu} & 57.50 & 502.49 & 10. & 10.0 & 22.2 \\ 
    & \mednextmkt{} \cite{Roy2023} & 32.65 & 248.03 & 15.32 & 67.6 & 153.6\\ 
    & \mednextmkf{} \cite{Roy2023} & 34.75 & 308.01 & 18.85 & 218.3 & 416.9 \\ 
    \midrule
    \multirow{5}{*}{\rotatebox{90}{Transf.}} 
    & TransUNet \cite{chen2021transunet} & 96.07 & 88.91 & 16.25 & 26.5 & 73.9 \\
    & \cotr{} \cite{xie2021cotr} & 50.12 & 369.22 & 8.10 & 18.6 & 41.4\\
    & \unetr{} \cite{hatamizadeh2022unetr} & 92.49 & 75.76 & 15.29 & 15.4 & 39.5\\
    & \swinunetr{} \cite{Hatamizadeh2022swin} & 62.83 & 384.20 & 13.91 & 22.0 & 38.7\\ 
    & \nnformer \cite{zhou2023nnformer} & 150.50 & 213.41 & 9.73 & 8.2 & 20.6\\ 
    \midrule
    \multirow{6}{*}{\rotatebox{90}{Mamba}}
    & \umambabot{} \cite{Ma2024} & 41.95 & 156.32 & 13.55 & 22.0 & 54.2 \\
    & \umambaenc{} \cite{Ma2024} & 42.85 & 231.18 & 26.42 & 37.9 & 89.3 \\
    & \SegMambaSkip{}$\dagger$ & 62.36 & 486.92 & 29.26 & 12.6 & 93.5 \\
    & \SegMamba{}$\dagger$ & 61.49 & 480.90 & 25.61 & 12.7 & 99.6\\
    & \PanSegMamba{}$\dagger$ & 64.75 & 494.17 & 27.31 & 16.5 & 134.1\\
    & \MultiSegMamba{}$\dagger$ & 68.46 & 527.56 & 36.92 & 18.2 & 149.0\\
    \bottomrule
    \end{tabular}
    \label{tab:parameters}
    \vspace{-15pt}
\end{table}

\smallskip
\noindent \textbf{Results.} As shown in \cref{tab:brain_results}, \cref{tab:synapse_results}, and \cref{tab:acdc_results} our proposed models consistently outperform all competing approaches, demonstrating superior overall performance across all the considered datasets. In general, \PanSegMamba{}, and \MultiSegMamba{} consistently outperform \SegMambaSkip{}, even if the latter is always competitive with state-of-the-art models and on some specific classes proves to be superior to all. 

As the results on the BrainTumor dataset (\Cref{tab:brain_results}) show, \SegMamba{}, \PanSegMamba{}, and \MultiSegMamba{} always outperform \SegMambaSkip{} on average metrics and on most individual classes taken separately. 
Among the SegMamba models, \MultiSegMamba{}, the one that harnesses more directions, outperforms the other configurations, demonstrating the importance of modeling multiple directions. Excluding \nnformer{}, our Mamba-based architectures gain more than 3 dice points over best performing transformer-based architectures and up to 1 dice point over \nnunet{}.

For what concerns the Synapse Abdomen dataset (\Cref{tab:synapse_results}), characterized by a larger number of classes, results show that our model showcase substantial improvements in kidney and spleen segmentation, as well as on average HD95 and DSC, when compared to state-of-the-art architectures. Remarkably, the inclusion of four distinct directions yields a more pronounced improvement on gallbladder segmentation, which is the most difficult to segment. Indeed, the gallbladder is significantly smaller and varies more in shape and position compared to other organs such as the liver, which is larger and more consistently shaped. Moreover, the close proximity of gallbladder to other organs and structures in the abdominal cavity increases the complexity of distinguishing it in medical images. 
Results on gallbladder segmentation show that \SegMamba{} reaches 62.21 Dice points, while its multidirection versions such as \PanSegMamba{} and \MultiSegMamba{} improve over it by 8 and 10 points respectively.

\begin{figure*}[tb]
    \centering
    \includegraphics[width=0.93\textwidth,valign=c]{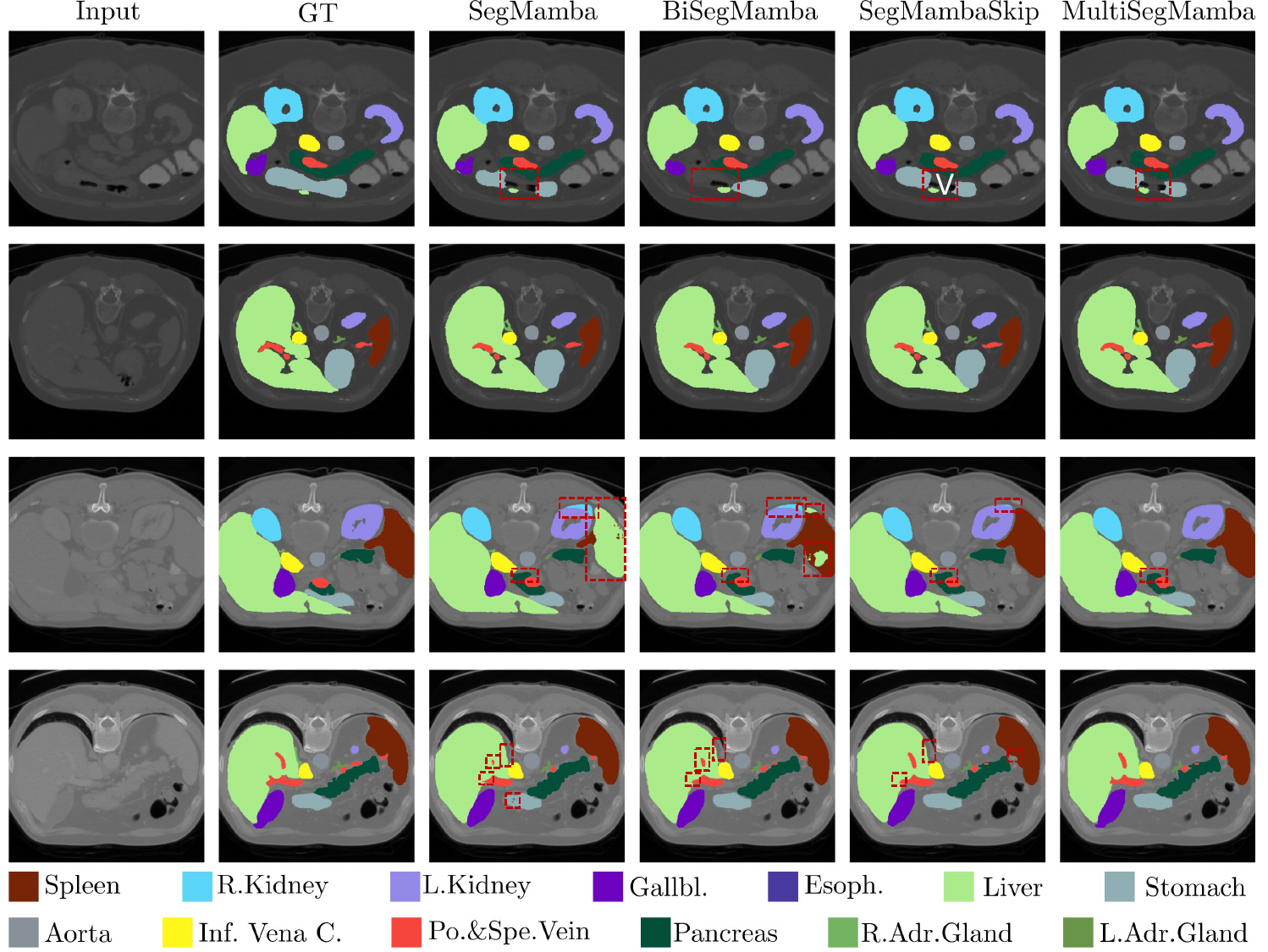}\label{fig:abdomen}
    \caption{
    Visualization of segmentation results for four sample cases from the Synapse Abdomen evaluation set. Annotation errors are marked with red dashed boxes. The figure is best viewed in color and zoomed in. From left to right: Input, Ground Truth (GT), \SegMamba{}, \PanSegMamba{}, \SegMambaSkip{}, and \MultiSegMamba{}.
    }
    \label{fig:qualitativa}
    \vspace{-10pt}
\end{figure*}
    
Finally, results on the ACDC dataset are presented in \Cref{tab:acdc_results}. This table highlights that  \MultiSegMamba{} outperforms all its variants that model less directions, and that \MultiSegMamba{} and \PanSegMamba{} consistently outperform \SegMambaSkip{} and all the U-Mamba variations.

\smallskip
\noindent \textbf{About Model Size.}
In \Cref{tab:parameters} a comprehensive computational comparison on the Synapse dataset is reported considering the number of parameters (millions), GFLOPs, and GPU memory. Our proposed models have a higher number of parameters compared to classical CNN approaches, while being comparable to or often having fewer parameters than transformer-based models. More specifically, the number of parameters of our models ($\sim$60M) are, on average, the double with respect to \nnunet{} ($\sim$31M), comparable to those of \nnunetres{} ($\sim$57M), and much lower than those of transformer-based models (from $\sim$95M of \transunet{} and \unetr{}, up to 150M of \nnformer{}). 

\smallskip
\noindent \textbf{Qualitative Evaluation.} \cref{fig:qualitativa} depicts a qualitative comparison of the four variations of the proposed architecture. The comparison is performed on samples taken from the Synapse Abdomen evaluation set. As can be seen, all of our Mamba-based variations perform qualitatively similarly, but the ones leveraging multiple directions are less prone to errors when dealing with fine-grained details. This confirms the quantitative results previously discussed.



\section{Conclusion}
\label{sec:conclusion}
This paper aims to assess the efficacy of the Mamba State Space Model for 3D medical image segmentation, comparing it with advanced convolutional and Transformer-based architectures. Additionally, we propose alternative designs for Mamba architectures to address their key limitations. Specifically, we integrate Mamba at various stages within the standard U-Net framework, either in skip connections or prior to pooling operations, utilizing both single-directional, bi-directional, and multi-directional implementations. The overall framework blends Convolutions and State Space Models, leveraging the former for encoding precise spatial information, while addressing the latter to model long-range voxel-level interactions. Mamba offer a dual advantage, providing a global context alongside voxel-wise precision, the former absent in traditional convolutional layers due to limited receptive fields and the latter absent in Transformers due to their computational complexity.

Our experimental results highlight the substantial improvement in HD95 and DSC metrics on three well-known datasets compared to \nnunet{} and different transformer-based networks. 
We showcase Mamba versatility by adapting it from its original use in text generation and large language models to achieve state-of-the-art results in a completely different task. This adaptability highlights Mamba potential beyond its initial design, demonstrating its efficacy on image encoding and segmentation.

\smallskip
\noindent \textbf{Limitations And Future Works.}
Despite the advancements made with the Mamba model, two key limitations can be identified. 

First, as Mamba is inherently a causal model, its application to non-causal visual data requires modification. Specifically, we tried to solve this problem by processing each sequence both forward and backward. Anyway, this introduces redundancy increasing the risk of overfitting. We believe that more efficient approaches could be developed to address this issue. 

Second, to capture spatial relationships, we unfold image patches from multiple directions, but more effective methods, such as identifying optimal scanning paths or partitioning larger volumes into smaller neighborhoods, may exist. Furthermore, employing too many directions can significantly increase computational demands and redundancy as mentioned before.

{\small
\balance
\bibliographystyle{ieee_fullname}
\bibliography{bibliography_mamba}
}

\end{document}


\title{Supplementary Material for: ``Taming Mambas for Voxel Level 3D Medical Image Segmentation''}
%
%
\author{}
%
\authorrunning{}
\titlerunning{}
%
\institute{}

%
\maketitle              
\vspace{-50pt}
\begin{figure*}[!h]
    \centering
    \includegraphics[width=0.8\textwidth,valign=c]{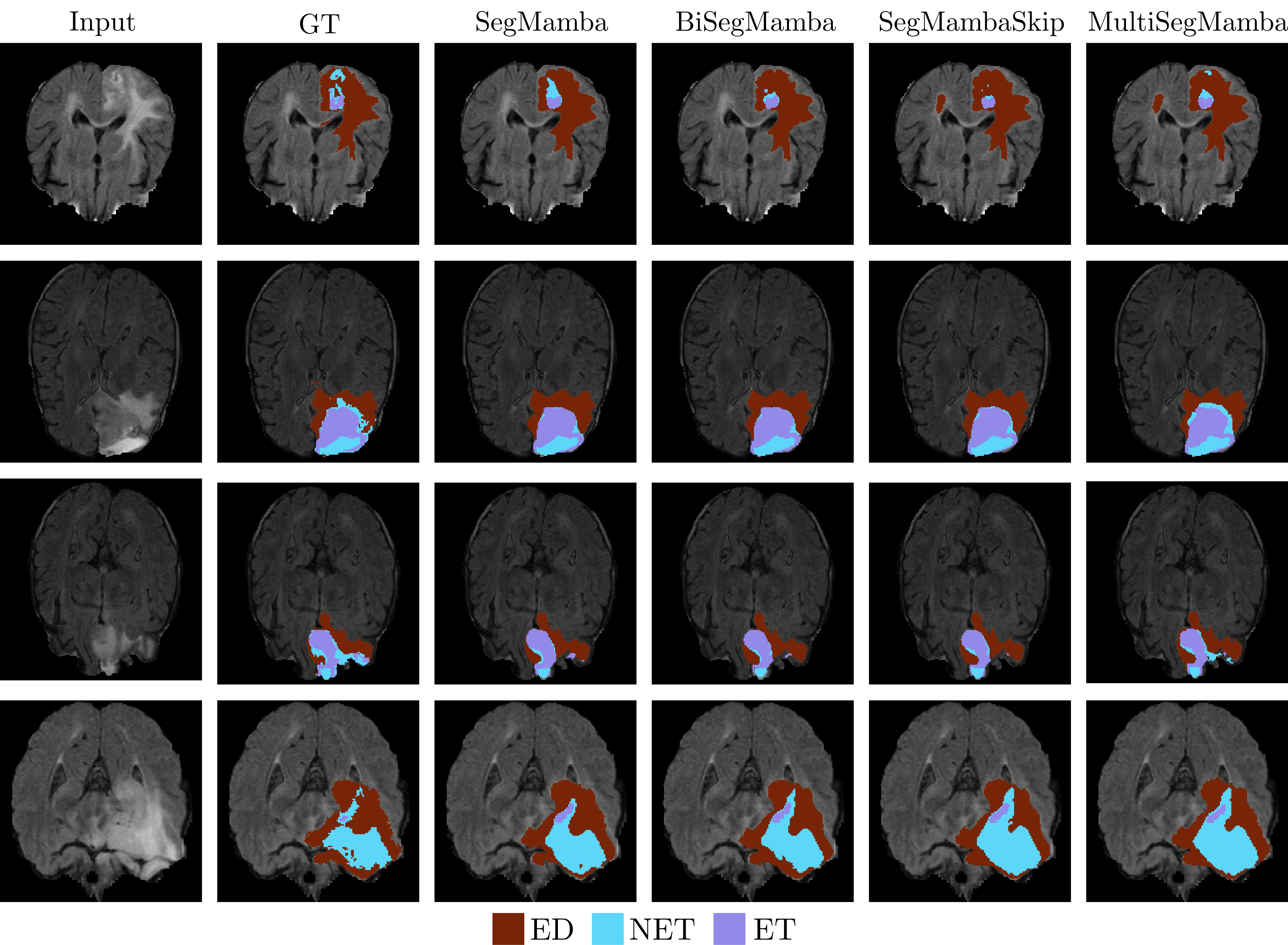}\label{fig:brain}
    \caption{Visualization of segmentation results on four sample cases taken from the BrainTumor dataset test set.}
\end{figure*}
\vspace{-10pt}
\begin{figure*}[!h]
    \centering
    \includegraphics[width=0.8\textwidth,valign=c]{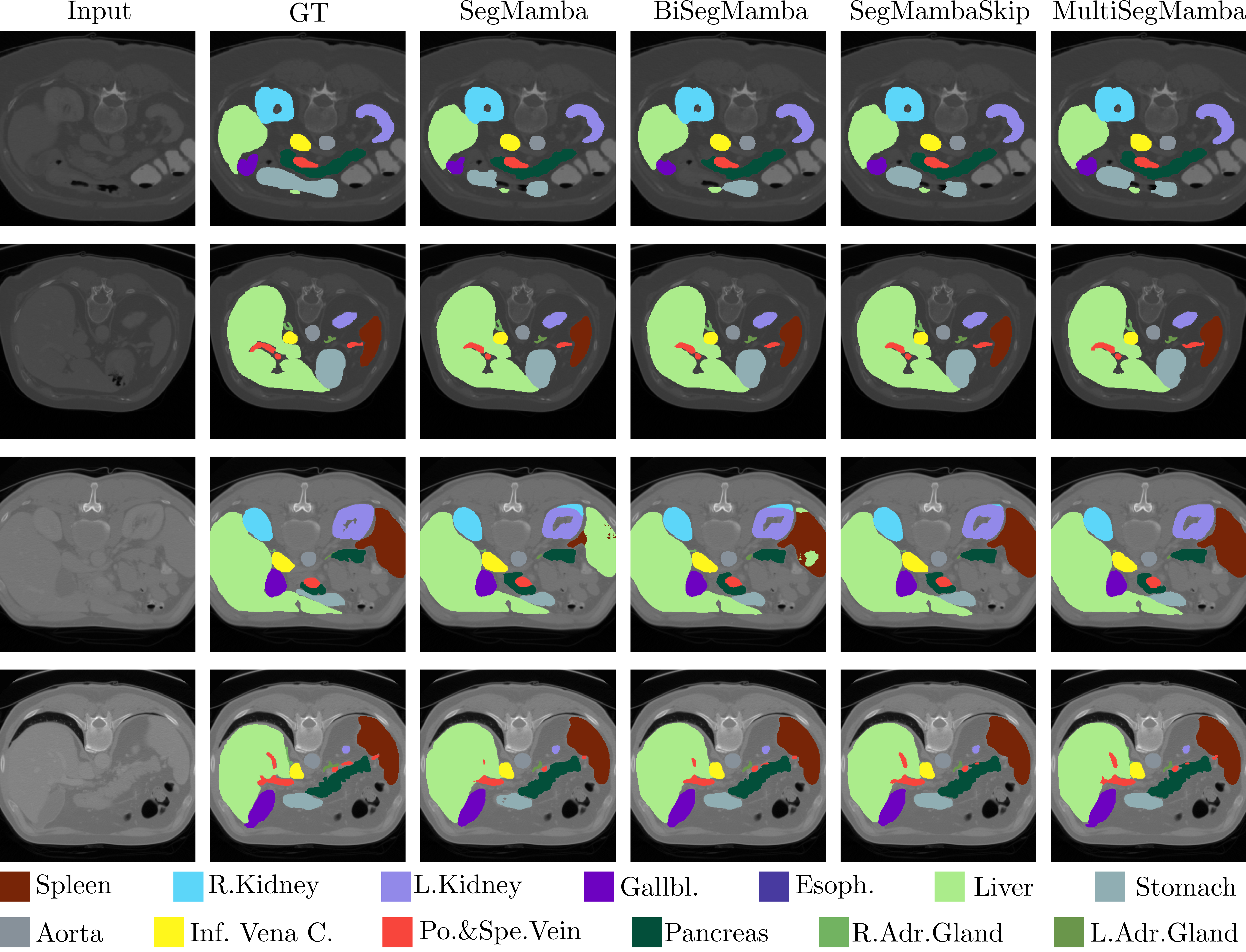}\label{fig:abdomen}
    \caption{Visualization of segmentation results on four sample cases taken from the Synapse Abdomen dataset test set.}
\end{figure*}

\begin{table}[t]
  \setlength{\tabcolsep}{6pt}
  \footnotesize
  \centering
  \caption{Configuration of the proposed models when trained on the selected datasets.}
  \setlength{\tabcolsep}{2pt} 
  \begin{tabular}{l
  >{\columncolor{PrimaryColumnColor}}c
  c
  >{\columncolor{PrimaryColumnColor}}c
  }
    \toprule
    & \cellcolor{SecondaryColumnColor} \textbf{BrainTumour} & \cellcolor{SecondaryColumnColor} \textbf{Synapse} & \cellcolor{SecondaryColumnColor} \textbf{ACDC} \\
    \midrule
    Spacing & [1, 1, 1] & [3, 0.76, 0.76] & [6.35, 1.52, 1.52] \\
    Median shape & $138 \times 170 \times 138$ & $148 \times 512 \times 512$ & $13 \times 246 \times 213$ \\
    Crop size    & $128 \times 128 \times 128$ & $48 \times 192 \times 192$ & $14 \times 256 \times 224$  \\
    Batch size   & 2 & 2 & 4 \\
    \bottomrule
  \end{tabular}
  \label{tab:config}
  \vspace{-10pt}
\end{table}

\begin{figure*}[!h]
    \centering
    \subfloat[]{\includegraphics[width=0.8\textwidth,valign=c]{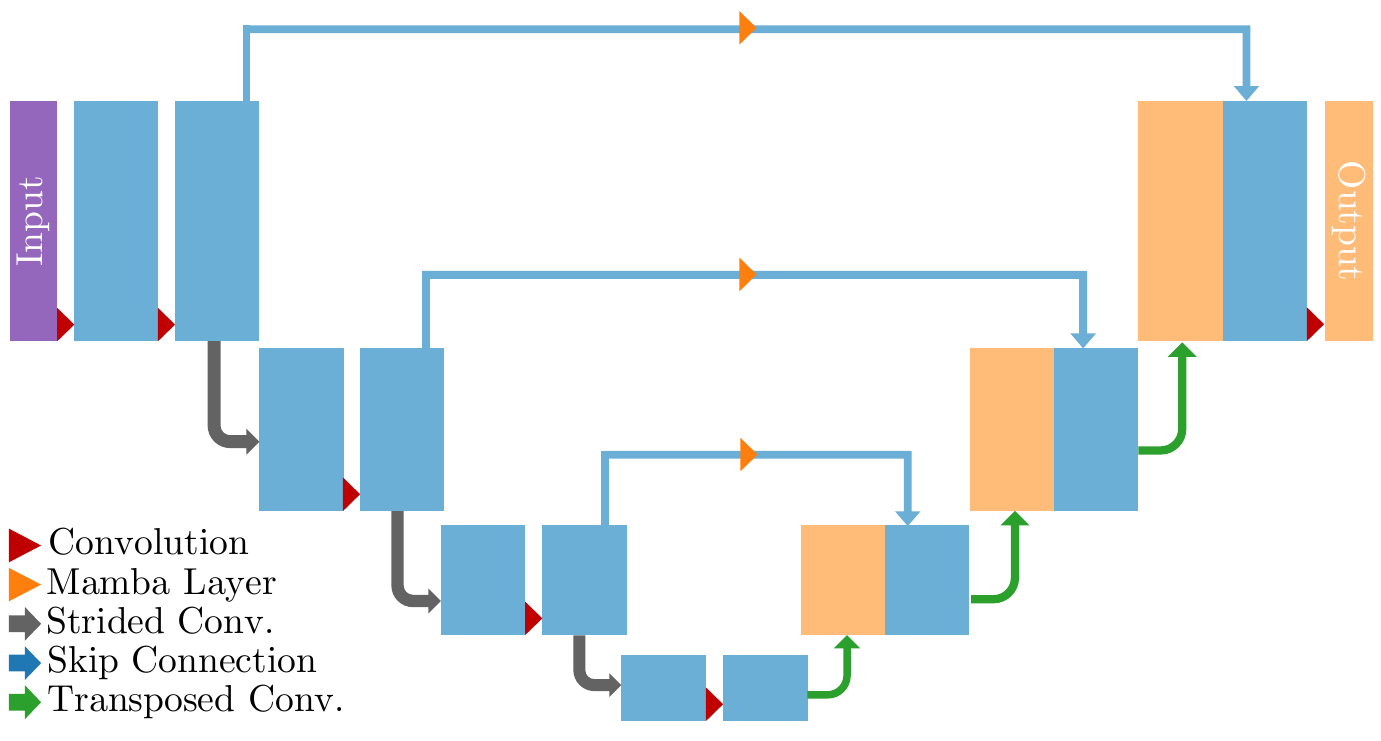}\label{fig:model}} \\
    \subfloat[]{\includegraphics[width=0.8\textwidth,valign=c]{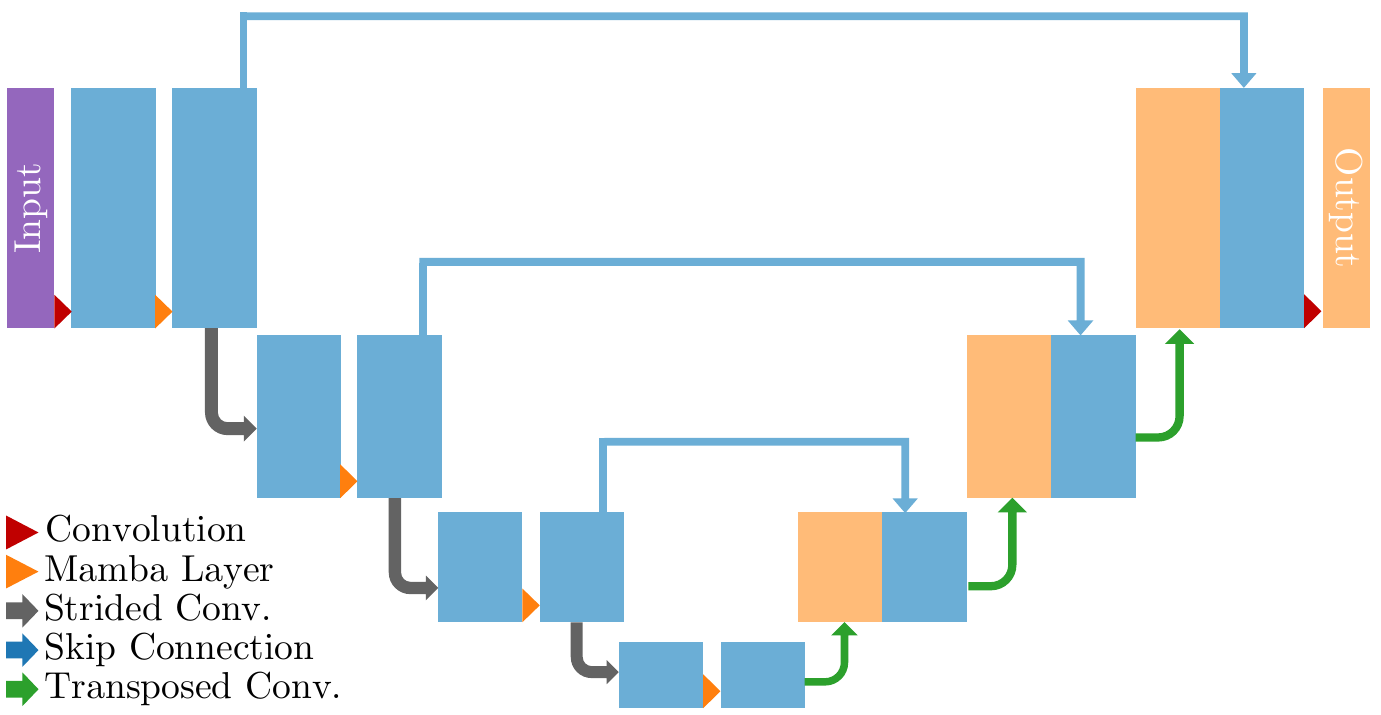}\label{fig:model}}
    \caption{In this figure the position of the proposed Mamba Layer(s) is depicted within the U-Net architecture. (a) correspond to \SegMambaSkip{}, (b) corresponds to all the others (\SegMamba{}, \PanSegMamba{}, \MultiSegMamba{}). }
\end{figure*}

\begin{table}[tb]
  \setlength{\tabcolsep}{6pt}
  \footnotesize
  \centering
  \caption{5-fold cross-validation results on the Automatic Cardiac Diagnosis (ACDC) dataset. Our proposals are marked with $\dagger$. The evaluation metric is the DSC (\%). Best results are in bold while second best are underlined.}
  \begin{tabular}{c l
  >{\columncolor{PrimaryColumnColor}}S[table-auto-round,table-format=2.2]
  c
  >{\columncolor{PrimaryColumnColor}}S[table-auto-round,table-format=2.2]
  c
  }
    \toprule
    & \textbf{Model} & \cellcolor{SecondaryColumnColor} \textbf{Average} & \textbf{RV} & \cellcolor{SecondaryColumnColor} \textbf{Myo}   & \textbf{LV}    \\
    \midrule
    \multirow{4}{*}{\rotatebox{90}{CNNs}} & \nnunet{} \cite{isensee2021nnu}  &91.42&90.1&88.74&95.41\\
    & \nnunetres \cite{isensee2021nnu} &90.84&89.17&88.52&94.84\\
    & \mednextmkt{} \cite{Roy2023} &91.64&89.43&\Uline{89.77}&95.72\\
    & \mednextmkf{} \cite{Roy2023} &90.7&88.5&88.88&94.73\\
    \midrule
    \multirow{9}{*}{\rotatebox{90}{Transformers}} & TransUNet \cite{chen2021transunet} &89.75&88.88&84.66&95.7\\
    & \transbts \cite{wang2021transbts} &91.29&90.42&87.94&95.51\\
    & \cotr \cite{xie2021cotr} &90.9&89.17&88.34&95.18\\
    & \unetr{} \cite{hatamizadeh2022unetr} &88.72&85.55&86.48&94.12\\
    & \swinunet{} \cite{cao2022swin} &89.97&88.29&85.61&\bfseries96.01\\
    & \swinunetr{} \cite{Hatamizadeh2022swin} &91.36&\Uline{90.48}&87.84&\Uline{95.75}\\
    & \levit{} \cite{xu2023levit} &90.21&89.78&87.1&93.75\\
    & \missformer{} \cite{huang2021missformer} &87.73&86.55&85.24&91.42\\
    & \nnformer{} \cite{zhou2023nnformer}    &\Uline{91.87}&\bfseries90.78&89.37&95.46\\
    \midrule
    \multirow{6}{*}{\rotatebox{90}{Mamba}}
    & \umambabot{} \cite{Ma2024} &90.44&87.67&88.76&94.89\\
    & \umambaenc{} \cite{Ma2024} &90.07&87.34&88.23&94.65\\
    & \SegMambaSkip{}$\dagger$ &91.49&89.58&89.51&95.39\\
    & \SegMamba{}$\dagger$  &91.33&89.37&89.4&95.22\\
    & \PanSegMamba{}$\dagger$  &91.5&89.46&89.66&95.37\\
    & \MultiSegMamba{}$\dagger$ &\bfseries 92.04&90.39&\bfseries90.29&95.44\\
    \bottomrule
  \end{tabular}
  \label{tab:acdc_results}
  \vspace{-0pt}
\end{table}